\documentclass{article}

\usepackage[final]{corl_2018} % Uncomment for the camera-ready ``final'' version
\usepackage{amsmath}
\usepackage{amsfonts}
\usepackage{amssymb}
\usepackage{graphicx}
\usepackage{algorithm}
\usepackage{algpseudocode}
\usepackage{tikz}
\usepackage{graphics} % for pdf, bitmapped graphics files
\usepackage{mathptmx} % assumes new font selection scheme installed
\usepackage{times} % assumes new font selection scheme installed
\usepackage{amsmath} % assumes amsmath package installed
\usepackage{mathtools}
\usepackage{amssymb}  % assumes amsmath package installed
\usepackage{amsfonts}
\usepackage{algorithm}
\usepackage{algpseudocode}

\usepackage{graphicx, wrapfig}
\usepackage{caption}
\usepackage{subcaption}
\usepackage{bm}
\usepackage{pgfplots}
\usepackage{pgfkeys}
\usepackage[export]{adjustbox}
\usepackage{color}
\usepackage[title]{appendix}

\title{Risk-Aware Active Inverse Reinforcement Learning}

\author{
  Daniel S. Brown$^*$, Yuchen Cui\thanks{these authors contributed equally}, and Scott Niekum\\
  Department of Computer Science\\
  University of Texas at Austin, United States\\
  \texttt{dsbrown@cs.utexas.edu, yuchencui@utexas.edu, sniekum@cs.utexas.edu} \vspace{-0.5cm}
}

\begin{document}
\maketitle

%===============================================================================

\begin{abstract}
Active learning from demonstration allows a robot to query a human for specific types of input to achieve efficient learning. Existing work has explored a variety of active query strategies; however, to our knowledge, none of these strategies directly minimize the performance risk of the policy the robot is learning. Utilizing recent advances in performance bounds for inverse reinforcement learning, we propose a risk-aware active inverse reinforcement learning algorithm that focuses active queries on areas of the state space with the potential for large generalization error. We show that risk-aware active learning outperforms standard active IRL approaches on gridworld, simulated driving, and table setting tasks, while also providing a performance-based stopping criterion that allows a robot to know when it has received enough demonstrations to safely perform a task.
\end{abstract}


%===============================================================================
\section{Introduction}
\label{sec:introduction}

With recent advancements in robotics, automation is expanding from factories
to homes, hotels, and hospitals. Given these less structured workplaces,
it is difficult to for engineers to anticipate the wide variety of tasks that a robot may be asked to perform. Learning from demonstration (LfD) \cite{argall2009survey} algorithms leverage human demonstrations, rather than code, to program robots, providing non-expert end-users with an intuitive way to program robots. 
Inverse Reinforcement Learning (IRL) \cite{ng2000algorithms,abbeel2004apprenticeship} is a form of LfD that assumes demonstrations are driven by an underlying reward function that captures the intention of the demonstrator. By recovering this reward function, the robot can perform policy improvement via reinforcement learning \cite{sutton1998reinforcement} and can generalize the task to novel settings.

However, it can be difficult for a human to understand what demonstrations would be most informative to provide to a robot, due to inherent physical, perceptual, and representational differences.
To address this issue, \textit{active} IRL algorithms \cite{lopes2009active,cohn2011comparing,cui2018active} have been proposed that
reason about uncertainty and information gain to select queries that are expected to be the most informative under certain criteria.
Existing active IRL algorithms aim to minimize uncertainty over policy \cite{lopes2009active}, minimize uncertainty
over possible reward functions \cite{cui2018active}, or maximize 
expected gain in policy value \cite{cohn2011comparing}. 
To allow robots to be programmed by non-experts, it is crucial that robots can reason about risk and generalization error given limited, ambiguous demonstrations. However, previous active IRL algorithms do not consider the risk of the actual policy learned by the robot. As an example of a demonstration that could lead to high generalization error, consider a human giving a single demonstration placing a vase on the center of a table. There are many plausible motivations for this demonstration: the vase should be in the center of table, the vase can be anywhere on the table, the vase should be above a cup, etc. Our paper seeks to addresses the following question: \textit{\textbf{How can a robot that learns from demonstrations actively query for help in states where its learned policy has the potential for high generalization error under the demonstrator's true reward function?}}  
%\scott{So far you've been talking about IRL, but this last sentence is phrased entirely in terms of policy learning.  Can you make a more explicit connection to IRL?}

%We propose a novel risk-aware active IRL algorithm that directly aims to minimize the risk of the robot's policy learned through demonstrations. Our approach queries for demonstrations in states where the robot has a chance of significant generalization error. 
%using the popular \textit{Value at Risk} (VaR) metric from finance \cite{jorion1997value}, which has been recently applied to bounding the performance of policies learned from demonstrations \cite{brown2018efficient} and can be especially beneficial for applications where safety is a concern. 
%\scott{There should be some discussion in the intro that ties this more closely to robotics and why the robotics community should care about safe IRL.  Give a concrete example or two.  So far, this is reading more like a NIPS paper than a CoRL paper.}

%The main contributions of our approach are: (1) (2) (3).

%===============================================================================
\section{Related Work}
\label{sec:related_work}
%\scott{This Related Work section seems like it is in the wrong place. This should come after the experiments section before the conclusion}
%\subsection{Risk-aware Learning}
There is a growing interest in making AI systems safe and well-behaved \cite{amodei2016concrete,thomas2017ensuring}. One common way to build safety into a system is to model risk. 
Risk-aware approaches have been applied to many different problems including planning \cite{chow2015risk,brown2017exact}, 
reinforcement learning \cite{tamar2015optimizing,garcia2015comprehensive}, and inverse reinforcement learning \cite{brown2018efficient}.
However, risk-aware active learning from demonstrations with high-confidence performance bounds has not been previously addressed.
%\textbf{TODO: cite that Abbeel and Moldorov? paper from a while back}

%\subsection{Learning from human feedback}
%Cite tamer and coach, maybe that paper from AAAI that was a blend of tamer and IRL. Talk about how they don't do active queries.
Leveraging interactive human feedback is an efficient way of learning desired robot behavior and has been explored widely in the literature. The TAMER framework proposed by \citet{knox2013teaching} and the COACH framework proposed by \citet{macglashan2017interactive} both cast interactive learning as a reinforcement learning (RL) problem. TAMER interprets human feedback as uniform reward signals while COACH argues that human feedback is policy dependent and should be treated as advantage signals. Under an RL paradigm, the policy learning agent is purely exploitative, and does not actively infer the reward function, leading to limited generalization. 
Recent work on active preference queries\cite{sadigh2017active,christiano2017deep} approximate a human's true reward function by querying a user's preference between two different policy rollouts. %\citet{sadigh2017active} assume a reward function linear in known features and generate queries to actively reduce the uncertainty over the relative weights of these features. \citet{christiano2017deep} use neural networks to both learn features and the reward function from preferences. 
%\subsection{Active Learning}
%\citet{javdani2014near} show that for Bayesian active learning, what matters is not reducing uncertainty in general, but to reduce uncertainty about decisions that matter. In a similar sense we show that for active IRL reducing uncertainty is not as important as reducing what really matters: reducing performance loss.

%\subsection{Active Inverse Reinforcement Learning}
Most active IRL algorithms use a Bayesian approach to systematically addresses the ambiguity in reward learning \cite{cohn2011comparing, cui2018active, lopes2009active,sadigh2017active}. 
\citet{cohn2011comparing} look at action-wise myopic gain for selecting the active query in order to maximize its potential gain in value,
while \citet{cui2018active}
directly compute expected information gain over the distribution of reward functions per state-action pair to maximally reduce uncertainty over the reward distribution.
These two methods compute the expected gain in value or information by doing inference over each state-action pair, which is computationally expensive in high dimensional state spaces, and therefore are not practical for real-time robotic active learning tasks.
The work of \citet{lopes2009active} reasons about state-wise policy entropy over the posterior distribution of policies given demonstrations
and asks for demonstrations at states with the highest policy entropies.
The state-wise policy entropy is ignorant of the actual policy that will be used for evaluation
and does not take the values of states into account. Additionally, \citet{lopes2009active} do not provide a semantically meaningful stopping condition for the active learning process. 
%Our proposed method directly improves the evaluation policy by evaluating states with their potential loss in value under the evaluation policy and stops when the VaR of evaluation policy is bounded by the desired threshold. \scott{This benefit should be discussed when introducing our algorithm in the previous section} 
%They synthesize queries based on expected feature count differences
%\textbf{TODO: describe their approach}. 
By contrast, our approach to active learning focuses directly on optimizing the objective of learning---the generalization performance of the robot's learned policy. 
%\textbf{TODO: we could maybe compare activeVaR against the sadigh halfspace idea for setting a table. Hopefully show that activeVaR is better at reducing policy loss than halfspaces focused only on cutting away volume. For high dimensions it probably suffers same way Cakmak does. Also only works for linear combo of features. We could show that ours works for learning weights and widths of rbf reward so nonlinear in reward parameters.} 

Existing active IRL approaches do not explicitly use the performance of the robot's learned policy to generate queries. One of the primary reasons for this is that, until recently, practical methods for estimating the performance of a policy learned from demonstrations when the ground-truth reward function is unknown, have not existed. Both \citet{abbeel2004apprenticeship} and \citet{syed2008game} derive theoretical Hoeffding bounds on performance based on matching feature counts; however, these bounds are too loose to use in practice \cite{brown2018efficient}. Recently, \citet{brown2017toward,brown2018efficient} proposed a Value-at-Risk \cite{jorion1997value} approach based on Bayesian IRL that computes accurate, tight bounds on the performance loss of any policy with respect to the optimal policy under the demonstrator's true, but unknown, reward function, but did not consider the case of active IRL.

%\textbf{Cite Singh's repeated IRL paper and Anca's active paper Scott sent}

%Our framework casts interactive learning as an IRL problem and generates active queries to expedite the learning process. Not only does our algorithm benefit from the generalizability of learned reward functions (instead of policies), but also our algorithm balances exploration and exploitation through active queries aimed at minimizing performance loss. \scott{Again, related work sections should mostly focus on the related work, not our algo.  This belongs in the earlier section.}

%===============================================================================
\section{Background Information}
\label{sec:background}
We model our problem as a Markov Decision Process and use a standard Bayesian approach to infer the reward function of the demonstrator.

\subsection{Markov Decision Processes}
A Markov Decision Process (MDP) is given by the tuple  $\langle S,A,T,R,d_0,\gamma \rangle$, where:
 $S$ is a set of states; $A$ is a set of actions; $T: S \times A \times S \to [0,1] $ is a transition probability function;
 $R: S \to \mathbb{R}$ is a reward function, with absolute value bounded by $R_{max}$;
$d_0$ is a starting state distribution and $\gamma \in [0,1)$ is the discount factor.
A policy, $\pi$ maps from states to a probability distribution over actions. The expected value of a policy $\pi$ under reward function $R$ is the expected return of that policy and is denoted as 
\begin{equation}
V^\pi_R = \mathbb{E}_{s_0 \sim d_0}[\sum_{t=0}^\infty \gamma^t R(s_t) | \pi].
\end{equation}
Similarly, the expected value of executing policy $\pi$ starting at any particular state $s \in S$ is defined as 
\begin{equation} V_R^{\pi}(s) = \mathbb{E}[\sum^{\infty}_{t=0}\gamma^t R(s_t) | s_0 = s , \pi]. \end{equation}
The Q-function is defined to describe values of state-action pairs according to some policy:
\begin{equation} Q^{\pi}(s,a) = R(s) + \gamma \mathbb{E}_{s'\sim T(s,a,\cdot)}[V^{\pi}(s')] \end{equation}

%Bellman equations are used to describe a recursive relationship between values of
%neighboring states and state-action pairs:
%\begin{equation}V^{\pi}(s) = R(s) + \gamma \sum_{s'}T(s,\pi(s),s')V^{\pi}(s') \end{equation}
%\begin{equation}Q^{\pi}(s,a) = R(s) + \gamma \sum_{s'}T(s,a,s')V^{\pi}(s') \end{equation}
%A policy $\pi$ is optimal if and only if:
%\begin{equation}\forall s \in S, \pi(s) \in \arg \max_{a \in A}Q^{\pi}(s,a)\end{equation}
%As commonly assumed in the literature \cite{abbeel2004apprenticeship,ziebart2008maximum,sadigh2017active}, the reward function is expressed as a linear combination of features:
%\begin{equation} R(s) = \omega^T \phi(s) \end{equation}
%\noindent where $\omega$ is a vector of feature weights. Therefore,
%the value of a policy $\pi$ given a reward function $R$ can be expressed as:
%\begin{equation} V^\pi_R = \mathbb{E}_{s_0\sim S_0}[\sum^\infty_{t=0}\gamma^t\omega^T\phi(s_t)|\pi] \end{equation}

\subsection{Bayesian Inverse Reinforcement Learning}

%Inverse reinforcement learning is an ill-posed problem since there are many reward functions that generate the same optimal policy \cite{ng2000algorithms}. 
We use Bayesian IRL \cite{ramachandran2007bayesian} to find the posterior distribution, $P(R|D) \propto P(D|R) P(R)$, over reward functions $R$ given demonstrations $D$.
%\begin{equation}
%P(R|D) \propto P(D|R) P(R).
%\end{equation}
%approaches this ill-posed problem in a principled way. In the formulation of BIRL \cite{ramachandran2007bayesian},
%we consider an MDP without reward function, denoted as MDP/R,
%$(S,A,T,d_0, \gamma)$ and an expert $\chi$ operating in the MDP. The expert $\chi$ is
%attempting to maximize the total accumulated reward according to some reward function $R^*$,
%using some stationary policy. The IRL agent observes
%a set of demonstrations $D = \{(s_0,a_0),(s_1,a_1)...(s_k,a_k)\}$. Since the policy used by $\chi$ is
%stationary, we can make the assumption that all the state-action pairs are independent evidence of some reward $R$: 
%\begin{equation}Pr(D|R) = \prod _{i=0}^{k} Pr((s_i,a_i|R)) \end{equation}
%According to equation (5), the reward-maximizing actions are equivalent to the actions with
%highest Q-values. Therefore, the likelihood of an action $(s_i,a_i)$ given a reward function $R$ can be modeled as:
%\begin{equation}Pr((s_i,a_i)|R) = \frac{1}{Z_i}e^{c Q(s_i,a_i,R)}\end{equation}
Bayesian IRL makes the common assumption that the demonstrator is following a softmax policy, resulting in the following likelihood function:
\begin{equation}
P(D|R) = \prod_{(s,a) \in D} \frac{e^{c Q^*_R(s,a)} }{\sum_{b \in A} e^{c Q^*_R(s,b)}}
\end{equation}
where $c$ is a parameter representing the degree of confidence we have in the demonstrator's ability
to choose the optimal actions \cite{ramachandran2007bayesian}, and $Q^*_R$ denotes the Q-function of the optimal policy under reward $R$. We use a uniform prior $P(R)$; however, other priors maybe used to inject domain specific or demonstrator specific knowledge.
%With Bayes theorem, the posterior probability of reward function $R$ is:
%\begin{equation}Pr(R|D) = \frac{Pr(D|R)Pr(R)}{Pr(D)} 
%{= \frac{1}{Z'}e^{c \sum_i Q(s_i,a_i,R)} Pr(R)}
%\end{equation} 
%While the normalizing constant $Z'$ is hard to compute, the Markov Chain Monte Carlo (MCMC)
%sampling algorithm only needs the ratios of probability densities and outputs
We use Markov Chain Monte Carlo (MCMC) sampling to obtain
a set of likely reward functions given the demonstrations, from which we can extract an estimate of the maximum a posteriori (MAP) reward function $R_{\text{MAP}}$ or the mean reward function $\bar R$.

%===============================================================================
\section{Methodology}
\label{sec:methodology}
We propose a general framework for risk-aware active queries based on the Value-at-Risk policy loss bounds for LfD proposed by \citet{brown2018efficient}. 
Our approach generates active queries that seek to minimize the policy loss Value-at-Risk of the robot's learned policy.

\subsection{Bounding the Performance of a Policy Given Demonstrations}
\label{subsec:boundingPolicyLoss}
As mentioned in the related work, existing active IRL approaches do not explicitly use performance as a query strategy. By applying and extending the work of \citet{brown2018efficient}, we demonstrate that it is possible to use performance-based active learning to improve a policy learned purely from demonstration. 

\citet{brown2018efficient} use samples from $P(R|D)$ to compute the policy loss $\alpha$-\textit{Value at Risk} ($\alpha$-quantile worst-case outcome) \cite{jorion1997value}.
Given an MDP$\setminus$R, a policy to evaluate $\pi_{eval}$, and a set of demonstrations $D$, the policy loss $\alpha$-VaR provides a high-confidence
upper bound on the $\alpha$-worst-case policy loss incurred by using $\pi_{eval}$ instead of $\pi^*$, where $\pi^*$ is the
optimal policy under the demonstrator's true reward function $R$.
The policy loss of executing $\pi_{\rm eval}$ under the reward $R$ is given by the Expected Value Difference: 
\begin{equation} \label{eq:EVDpolicy}
\text{EVD}(\pi_{eval},R) = V^{\pi^*}_{R} - V^{\pi_{eval}}_{R}.
\end{equation}

However, IRL is ill-posed---there are an infinite number of reward functions that explain any optimal policy \cite{ng2000algorithms}. Thus, any method that attempts to bound the performance of a policy given only demonstrations should account for the fact that there is never one ``true'' reward function, but rather a set of reward functions that motivate a demonstration.
%Therefore, VaR is a $(1-\delta)$ confidence upper bound on $\nu_{\alpha}(\text{EVD}(\pi_{eval},R^*))$, where $R \sim Pr(R|D)$.

To bound the policy loss $\alpha$-VaR of an evaluation policy $\pi_{\rm eval}$ given only demonstrations, \citet{brown2018efficient} propose using Bayesian IRL to generate samples of likely reward functions, $R$, from the posterior $P(R|D)$. Each sample reward function produces a sample policy loss, $\text{EVD}(\pi_{eval},R)$, and these policy losses can then be sorted to obtain a single sided $(1 - \delta)$-confidence bound on the policy loss $\alpha$-VaR \cite{brown2018efficient}. Note that this method uses Markov Chain Monte Carlo simulation to estimate the Value-at-Risk. This allows estimation of the Value-at-Risk over any distribution $P(R|D)$.

%For each $R_i \in R$, we can calculate:
%\begin{equation} Z_i = \text{EVD}(\pi_{eval},R_i) = V^*_{R_i} - V^{\pi_{eval}}_{R_i}\end{equation}
%\noindent which gives us samples from the posterior distribution over expected value differences.
%These samples are then sorted in ascending order to find a single-sided $(1-\delta)$ confidence bound
%on the $\alpha$-VaR. For any sample $Y_j$ in the sorted list, we have:
%\begin{equation} Pr(\nu_\alpha(Z) \leq Y_j) = \sum_{i=1}^j \binom{N}{i} \alpha^i (1-\alpha)^{N-i}  \\
%  \approx F_{\mathcal{N}}\big(j + \frac{1}{2} | N\alpha, N\alpha(1-\alpha)\big) \end{equation}
%To find the $k$ such that $ Pr(\nu_\alpha (Z) \leq Y_k) \geq (1-\delta) $, we can invert equation (15) and get:
%\begin{equation} k = [N\alpha + F^{-1}_{\mathcal{N}}(1-\delta)\sqrt{N\alpha(1-\alpha)} - \frac{1}{2}]\end{equation}

\subsection{Risk-Aware Active Queries}
The method proposed by \citet{brown2018efficient} works for any evaluation policy $\pi_{\rm eval}$. In an active learning setting, we argue that the most useful policy to evaluate is the robot’s best guess of the optimal policy under the demonstrator’s reward. Thus, we use $\pi_{\rm eval} = \pi_{\rm MAP}$ where $\pi_{\rm MAP}$ is the optimal policy corresponding to $R_{\rm MAP}$, the maximum a posteriori reward given the demonstrations so far; however, our general approach can easily be applied to any policy learned from demonstrations.

Rather than directly using the approach described in Section~\ref{subsec:boundingPolicyLoss} to bound the expected $\alpha$-Value-at-Risk policy loss of the policy $\pi_{\text{MAP}}$,
we instead generate risk-aware active queries by calculating the $\alpha$-VaR for each potential query state $s$ and select the state with the highest policy loss $\alpha$-VaR as the query. 
The policy loss of a state $s$ under $\pi_{\text{MAP}}$ given reward $R$ is:
\begin{equation} \label{eq:EVDstate}
Z = \text{EVD}(s,R \mid \pi_{\rm MAP}) = V^{\pi^*}_{R}(s)-V^{\pi_{\text{MAP}}}_{R}(s)
\end{equation}
where $\pi^*$ denotes the optimal policy under $R$.
Therefore, the $\alpha$-VaR for a state can be found using an extension of the method discussed in Section~\ref{subsec:boundingPolicyLoss}. We first sample rewards $R\sim P(R|D)$ using Bayesian IRL \cite{ramachandran2007bayesian}, then for each sampled $R$ and each potential query state $s$, we compute a sample policy loss $Z$ using Equation~\ref{eq:EVDstate}. These samples for each state can then be used to obtain one-sided confidence bounds on the $\alpha$-VaR for each state \cite{brown2018efficient}. Note that the state value functions $V^{\pi^*}_{R}(s)$ can be efficiently computed using the optimal Q-value functions computed during Bayesian IRL.

We consider two types of interactions with the demonstrator: active action queries and active critique queries. An active action query is where the robot proposes a state as its query and a human demonstrator is expected to provide the correct action to take at that state. The active learning process is summarized in Algorithm~\ref{alg:activeVaR}. 
%\scott{You should provide some intuition and/or justification here about why you are greedily selecting the argmax over states at each iteration.  Why is this a good approximation of your actual objective -- joint minimization of VaR with the fewest number of queries (you should also mention this objective somewhere if you haven't)? It would also be good to have a bit of exposition describing the algorithms instead of just the pseudocode.}
The objective of our active learning algorithm is to minimize the worst-case generalization error of the learned policy, with as few queries as possible. To do this we compute the $\alpha$-VaR for each starting state state (sampling states for continuous environments). The state with the highest $\alpha$-VaR under $P(R \mid D)$ is the state where the robot's policy has the highest $\alpha$-quantile worst-case policy loss. Thus, selecting this state as the active query is a good approximation for our objective.

\begin{algorithm}[t]
%\floatconts
{\caption{Action Query ActiveVaR( Input: MDP$\setminus$R, $D$, $\alpha$, $\epsilon$; Output:  $R_{MAP}$, $\pi_{MAP}$)} \label{alg:activeVaR}}
{
%\begin{algorithmic}[1] 
\begin{enumerate}
  \item Sample a set of reward functions $R$ by running Bayeisan IRL with input $D$ and MDP$\setminus$R; 
  \item Extract the MAP estimate $R_{MAP}$ and compute $\pi_{MAP}$;
  \item \textbf{while} \textit{true}:
  \begin{enumerate}
  \item $s_k = \arg \max_{s_i\in S} (\alpha\text{-VaR}(s_i,\pi_{MAP}))$ ;
  \item Ask for expert demonstration $a_k$ at $s_k$ and add $(s_k,a_k)$ into demonstration set $D$;
  \item Sample a new set of rewards $R$ by running Bayesian IRL with updated $D$; 
  \item Extract the MAP estimate $R_{MAP}$ and compute $\pi_{MAP}$;
  \item \textbf{break} if $\max_{s_i\in S} (\alpha\text{-VaR}(s_i,\pi_{MAP})) < \epsilon$;
  \end{enumerate}
  \item \textbf{return} $R_{MAP}$, $\pi_{MAP}$
\end{enumerate}
%\EndWhile \\
%\Return $R_{MAP}$, $\pi_{MAP}$; 
%\end{algorithmic} 
}

\end{algorithm}

In the second type of query, the robot demonstrates a trajectory to the user and asks them to critique the demonstration by segmenting the demonstration into desirable and undesirable segments as proposed by \citet{cui2018active}. To generate a trajectory for critique, the state with the highest $\alpha$-VaR policy loss is computed. A trajectory is then generated by executing the MAP policy starting at the selected state. This active learning process can be obtained from Algorithm~\ref{alg:activeVaR}, with the following replacements: (b) Ask for critique of the trajectory of length $L$ under $\pi_{MAP}$ starting at $s_k$ and (c) Update $D$ with positive and negative segments and sample new set of rewards with Bayesian IRL.

% \begin{algorithm}[htbp]
% %\floatconts
% {\caption{Critique Query ActiveVaR( Input: MDP/R, $D$, $c$, $\alpha$, $\delta$, $\epsilon$, $L$; Output:  $R_{MAP}$;)} \label{alg:activeVaR_traj} }
% {
% %\begin{algorithmic}[1] 
% \begin{enumerate}
%   \item Sample a set of reward functions $R$ by running BIRL with input $D$ and MDP/R; 
%   \item Extract the MAP estimate $R_{MAP}$ and compute $\pi_{MAP}$;
%   \item \textbf{while} \textit{true}:
%   \begin{enumerate}
%   \item $s_k = \arg \max_{s_i\in S} (VaR(s_i,\pi_{MAP}))$ ;
%   \item Find the trajectory $T_k$ of length $L$ under $\pi_{MAP}$ starting at $s_k$;
%   \item Ask for expert critiques on $T_k$ and obtain optimal and update $D$;
%   \item Sample a new set of rewards $R$ by running BIRL with updated demonstrations;
%   \item Extract the MAP estimate $R_{MAP}$ and compute $\pi_{MAP}$;
%   \item \textbf{return} if $\max_{s_i\in S} (VaR(s_i,\pi_{MAP})) < \epsilon$;
%   \end{enumerate}
% \end{enumerate}
% %\EndWhile \\
% %\Return $R_{MAP}$, $\pi_{MAP}$; 
% %\end{algorithmic} 
% }

% \end{algorithm}

Both of the above active learning algorithms repeat until the $\alpha$-VaR of the robot's learned policy falls below the desired safety threshold $\epsilon$. Further discussion of how to choose a meaningful stopping condition can be found in Appendix \ref{app:discussion}. We call our framework Risk-Aware Active IRL and refer to the corresponding active learning algorithms as \textit{ActiveVaR} in future discussion.\footnote{Our implementation of ActiveVaR can be found at: \url{https://github.com/Pearl-UTexas/ActiveVaR.git} }

\subsection{Example}
As discussed in \ref{sec:related_work}, given samples of reward functions from running Bayesian IRL, there are many different statistical measurements that can serve as the objective function for active learning \cite{cui2018active,lopes2009active,cohn2011comparing}. Many existing methods \cite{cui2018active,lopes2009active} are purely exploratory since they reason only about statistical measures instead of the performance of a specific policy when selecting active queries. Our proposed framework instead uses focused exploration based on policy loss.

\begin{figure}
\centering
\includegraphics[scale=0.23]{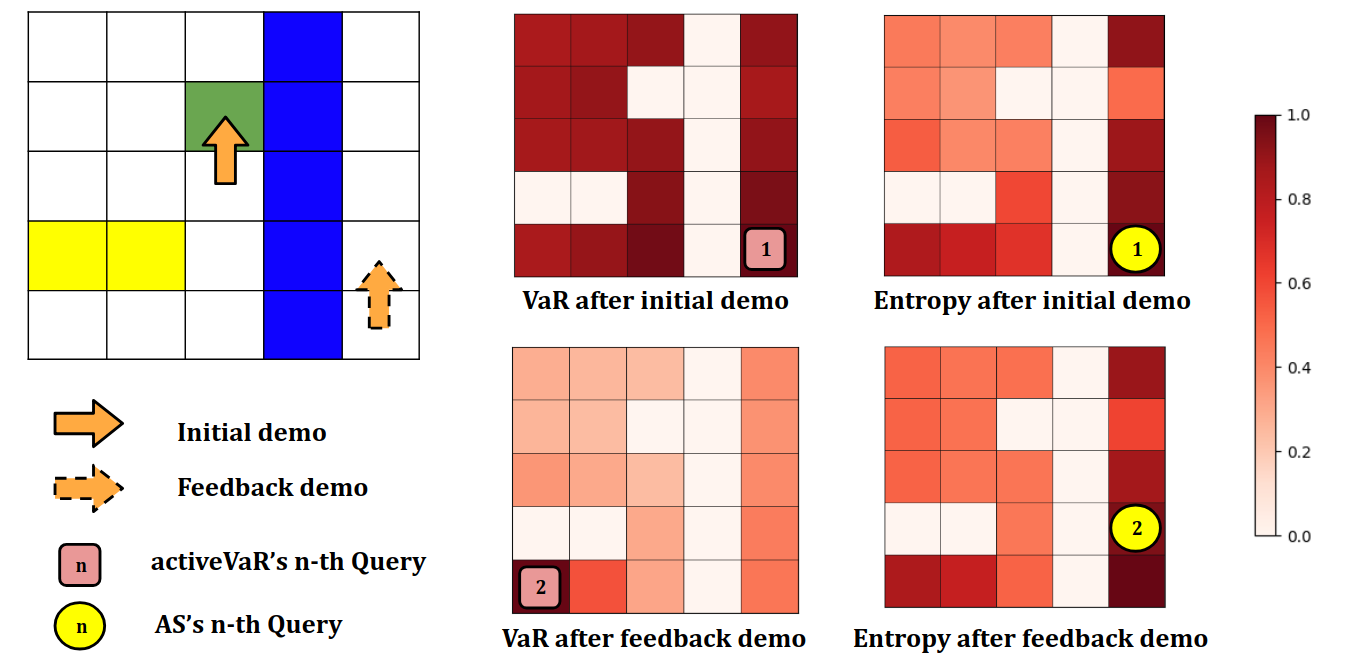}
\caption{\small Comparison of active action queries based on performance loss risk or action entropy. The example gridworld has four different unknown features denoted by the yellow, green, white, and blue colors of the cells. White states are legal initial states. AS is the active learning algorithm proposed by Lopes et al. \cite{lopes2009active} and activeVaR is our proposed active action query algorithm. The first two active queries proposed by each algorithm are annotated on the heatmaps of VaR and entropy values after each iteration. For heatmaps, all values are normalized from 0 to 1. }
\label{fig:intuitiveExample}
\end{figure}

The example in Figure~\ref{fig:intuitiveExample} shows how focusing on Value-at-Risk, rather than minimizing uncertainty over actions as in the Active Sampling (AS) algorithm \cite{lopes2009active}, leads to more intuitively intelligent queries. 
%When there are undesirable states the agent cannot avoid when starting at particular states, activeVaR takes it out of the picture, while AS may still think there is high entropy in these specific states because the weight for that feature hasn't been learned yet. 
In this example, there are four indicator features denoted by the white, yellow, green, and blue colors on the cells. Given one initial demonstration going into the green state from a white state, AS and ActiveVaR both pick the bottom right white state as their first query. However, for the second query, AS picks another state next to a blue feature while ActiveVaR picks the state next to a yellow feature.
Active queries based on VaR allow the IRL agent to understand that blue feature is unavoidable from all the rightmost states so there is no point asking for more demonstrations from those states while AS only reasons about action entropy. 
%\scott{Quickly? Or is it fundamentally not able to come to this conclusion at all, based on the way that it reasons?}

%===============================================================================
%===============================================================================
\section{Experiments}
\label{sec:experiments}

\subsection{Gridworld Active Action Queries}
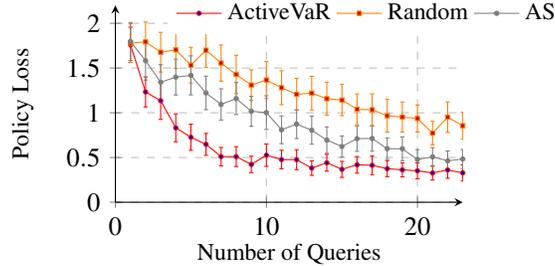
\begin{figure}
\centering
    \begin{tikzpicture}    
        \begin{axis}[
            name=plot1,
            xlabel={\small Number of Queries},
            x label style={at={(axis description cs:0.5,0.04)},anchor=north},
            ylabel={\small Policy Loss},
            axis x line=bottom,
            axis y line=left,
            width=6.2cm,height=4.2cm,
            xmin=0, xmax=23,
            ymin=0, ymax=2.2,
            ymajorgrids=true,
            xmajorgrids=true,
            xmajorgrids=true,
            grid style=dashed,
            legend style={at={(0.15,0.96)},anchor=west},
            legend columns=3,
            legend style={draw=none},
        ]
             \addplot+[
            color=red, 
               mark size=0.8,
            error bars/.cd,y dir=both,y explicit
            ]
            coordinates {
(1,1.7595636942675152) +- (0.19713236645458623,0.19713236645458623)
(2,1.2331263057324842) +- (0.1673687041421183,0.1673687041421183)
(3,1.134844012738854) +- (0.2088047014766918,0.2088047014766918)
(4,0.8319198089171979) +- (0.15570223057151428,0.15570223057151428)
(5,0.7263782165605093) +- (0.14610157891184233,0.14610157891184233)
(6,0.6465110191082805) +- (0.12270924026375282,0.12270924026375282)
(7,0.5098666878980892) +- (0.10308011867541927,0.10308011867541927)
(8,0.5097773885350319) +- (0.11126096403969186,0.11126096403969186)
(9,0.4227641401273887) +- (0.09138211359310021,0.09138211359310021)
(10,0.5256099999999996) +- (0.12520560526748534,0.12520560526748534)
(11,0.47689101910828) +- (0.10364593637123212,0.10364593637123212)
(12,0.47566448407643325) +- (0.11127784921094605,0.11127784921094605)
(13,0.38150577070063674) +- (0.0809317353774348,0.0809317353774348)
(14,0.4410903821656054) +- (0.09373834993194695,0.09373834993194695)
(15,0.36746936305732486) +- (0.08954544128074995,0.08954544128074995)
(16,0.41823461146496815) +- (0.09305799789168047,0.09305799789168047)
(17,0.4124921656050955) +- (0.10364228592012249,0.10364228592012249)
(18,0.3753649044585985) +- (0.08930081431649224,0.08930081431649224)
(19,0.3619137579617835) +- (0.07871161025235361,0.07871161025235361)
(20,0.3510034649681528) +- (0.0892224190411674,0.0892224190411674)
(21,0.32733771974522297) +- (0.07898663999197346,0.07898663999197346)
(22,0.3606938216560511) +- (0.09300181947118549,0.09300181947118549)
(23,0.328014675159236) +- (0.08998591708025307,0.08998591708025307)
(24,0.328346687898089) +- (0.07435333112022875,0.07435333112022875)
(25,0.3630025350318473) +- (0.08995478983126477,0.08995478983126477)

            };
            
    \addplot+[
            color=orange, 
            mark size=0.8,
            error bars/.cd,y dir=both,y explicit
            ]
            coordinates {
(1,1.7782213375796174) +- (0.21600311641648645,0.21600311641648645)
(2,1.7923898089171977) +- (0.22167195428940684,0.22167195428940684)
(3,1.677522420382166) +- (0.22166718903470306,0.22166718903470306)
(4,1.703330382165605) +- (0.2072616278027641,0.2072616278027641)
(5,1.532254140127389) +- (0.19944674519505817,0.19944674519505817)
(6,1.699758726114649) +- (0.19404213974286724,0.19404213974286724)
(7,1.5558385350318462) +- (0.2023669007281522,0.2023669007281522)
(8,1.4278561146496809) +- (0.21468926102766356,0.21468926102766356)
(9,1.3096474522292987) +- (0.17059875331113777,0.17059875331113777)
(10,1.3663464968152867) +- (0.20767923019467854,0.20767923019467854)
(11,1.280250382165605) +- (0.1895672384775975,0.1895672384775975)
(12,1.2065684076433119) +- (0.17764758759303217,0.17764758759303217)
(13,1.2201877070063694) +- (0.18783522817970866,0.18783522817970866)
(14,1.1569951592356698) +- (0.18338276400098363,0.18338276400098363)
(15,1.1423178343949052) +- (0.17424407390138266,0.17424407390138266)
(16,1.0392422929936305) +- (0.17168343634853617,0.17168343634853617)
(17,1.033768853503185) +- (0.17848500722121424,0.17848500722121424)
(18,0.9666324840764322) +- (0.18277907483805336,0.18277907483805336)
(19,0.9528762547770697) +- (0.1677253177226196,0.1677253177226196)
(20,0.9359508917197453) +- (0.15066371904123602,0.15066371904123602)
(21,0.7730421783439492) +- (0.13201040656906007,0.13201040656906007)
(22,0.9513682547770691) +- (0.17052980310665228,0.17052980310665228)
(23,0.8554114904458602) +- (0.15038874829942908,0.15038874829942908)
(24,0.7451650318471333) +- (0.14443686442965778,0.14443686442965778)
(25,0.7782434522292991) +- (0.15685420810625153,0.15685420810625153)

       }; 
            
             \addplot+[
        color=gray, 
           mark size=0.8,
        error bars/.cd,y dir=both,y explicit
        ]
        coordinates {
    (1,1.7969762420382176) +- (0.21234955944256506,0.21234955944256506)
(2,1.5820980891719751) +- (0.21652008932936948,0.21652008932936948)
(3,1.3414915286624212) +- (0.19637072671993377,0.19637072671993377)
(4,1.3989278980891733) +- (0.20010095919582901,0.20010095919582901)
(5,1.417984394904458) +- (0.21777810752932433,0.21777810752932433)
(6,1.2220331847133765) +- (0.18768396479419305,0.18768396479419305)
(7,1.0932421019108278) +- (0.17474058370769102,0.17474058370769102)
(8,1.159285796178344) +- (0.1787865983988154,0.1787865983988154)
(9,1.0169068789808908) +- (0.16663573650699076,0.16663573650699076)
(10,1.001993375796178) +- (0.18865172382175005,0.18865172382175005)
(11,0.8100636305732483) +- (0.15538597413232497,0.15538597413232497)
(12,0.8726548789808914) +- (0.160791930988377,0.160791930988377)
(13,0.8051129936305736) +- (0.15579050127791577,0.15579050127791577)
(14,0.6947454649681531) +- (0.14765800544266156,0.14765800544266156)
(15,0.6230002547770703) +- (0.11788425929417658,0.11788425929417658)
(16,0.7107886624203823) +- (0.1421422799401614,0.1421422799401614)
(17,0.7125334394904462) +- (0.14203493659806277,0.14203493659806277)
(18,0.5983207388535033) +- (0.11999092854549599,0.11999092854549599)
(19,0.5972540764331211) +- (0.1326003359337524,0.1326003359337524)
(20,0.47804057324840743) +- (0.1105964306965699,0.1105964306965699)
(21,0.5072876433121023) +- (0.1053020662728629,0.1053020662728629)
(22,0.46315249044585965) +- (0.11001847761873818,0.11001847761873818)
(23,0.4831113375796179) +- (0.1056946205807534,0.1056946205807534)
(24,0.5065876178343945) +- (0.11132083130433137,0.11132083130433137)
(25,0.42529747133757956) +- (0.10354587233177973,0.10354587233177973)

        }; 
          \addlegendentry{\small ActiveVaR}   
          \addlegendentry{\small Random}
                \addlegendentry{\small AS} 
\end{axis}
\end{tikzpicture}

\caption{\small Active action queries: averaged policy losses in 8$\times$8 gridworlds with 48 features.}
\label{fig:action_query}
\vspace{-0.2cm}
\end{figure}

We first evaluate ActiveVaR on active action queries in which the active learning algorithm selects a state and asks the demonstrator for an action label. The Active Sampling (AS) algorithm proposed by \citet{lopes2009active} can only work with action queries. AS is not as computationally efficient as activeVaR, but is much more computationally efficient than methods that require sampling hypothesis probability distributions \cite{cui2018active,cohn2011comparing}.
We conducted experiments in simulated gridworlds of size 8x8 with 48 random continuous features. The ground-truth feature weights are generated randomly and normalized such that the L1 norm of the weight vector is 1. The expected losses in return comparing to the optimal policy are measured and plotted in Figure~\ref{fig:action_query}, where \textit{Random} is a baseline that selects a random state as an active query per iteration. ActiveVaR is able to rapidly reduce the policy loss over iterations since it directly optimizes for performance.

\subsection{Gridworld Active Critique Queries}
The previous section demonstrated that action queries using a risk-aware active learning approach based on expected performance loss, or risk, outperforms standard entropy-based queries. We now demonstrate that our risk-aware active learning framework also allows active critique queries \cite{cui2018active}. \citet{cui2018active} proposed a novel active learning algorithm, ARC, where the robot actively chooses sample trajectories to show the human and the human critiques the trajectories by segmenting them into good and bad segments. 
This type of active query can be more natural for a human than giving an out of context action for a state and only requires the human to be able to recognize, not demonstrate, desirable and undesirable behavior. However, synthesizing trajectories for critique requires costly Bayesian inference over possible segmentations. Additionally, it is purely exploratory since it reasons only about information gain in the reward belief space without reasoning about the performance of current learned policy. %\textbf{Say something about how it doesn't use performance just KL divergence}.

To generate risk-aware performance-based trajectories we examine the robot's policy and evaluate per-state policy loss 0.95-VaR. Because the VaR is calculated based on the value of executing the robot's current policy from that state, we sample a trajectory from the robot's policy starting at that state and ask the demonstrator to critique it. We conducted experiments in 8x8 gridworlds with 48 different continuous features and allow each algorithm to generate a trajectory query of length 8 per critiquing iteration. As baselines, \textit{Random} selects an active query by rolling out the current MAP policy from a randomly selected state.  As shown in Figure~\ref{fig:traj_query}, while ActiveVaR's performance is not as good as that of ARC per number of trajectory queries, the time required to run ActiveVaR is much less than that of ARC. Because ARC's inference algorithm is sequential, not parallelizable, ActiveVaR is more practical for real-time active learning scenarios, while also outperforming random queries. Additional experiments were performed to highlight cases when ActiveVaR outperforms Random by a much larger margin (see Appendix \ref{app:results}).

%\textbf{@Yuchen: Do we have any results comparing activeVaR and your method when receiving critiques?} 

\begin{figure}

\centering
\begin{subfigure}[b]{0.45\linewidth}
    \begin{tikzpicture}           
        \begin{axis}[
            name=plot1,
            xlabel={\small Number of Trajectory Queries},
            x label style={at={(axis description cs:0.5,0.04)},anchor=north},
            ylabel={\small Policy Loss},
            axis x line=bottom,
            axis y line=left,
            width=5.5cm,height=4.2cm,
            xmin=0.8, xmax=12.5,
            ymin=0, ymax=4.55,
            ymajorgrids=true,
            xmajorgrids=true,
            y label style={at={(axis description cs:0.13,0.26)},anchor=west},
            xmajorgrids=true,
            grid style=dashed,
            legend style={at={(0.3,0.85)},anchor=west},
            legend columns=2,
            legend style={draw=none},
        ]
             \addplot+[
            color=red, 
               mark size=0.8,
            error bars/.cd,y dir=both,y explicit
            ]
            coordinates {
(1,3.282570930232557) +- (0.4035343297726813,0.4035343297726813)
(2,3.082726627906977) +- (0.5256856736287641,0.5256856736287641)
(3,2.216844651162791) +- (0.4167640905740301,0.4167640905740301)
(4,1.6215343720930235) +- (0.44580568608650434,0.44580568608650434)
(5,0.9712588372093027) +- (0.28163189486213314,0.28163189486213314)
(6,0.8815409883720929) +- (0.3338543515115426,0.3338543515115426)
(7,0.55363) +- (0.17962892470846598,0.17962892470846598)
(8,0.532377604651163) +- (0.21208592000264656,0.21208592000264656)
(9,0.2653510465116279) +- (0.14219089596659407,0.14219089596659407)
(10,0.18933636046511634) +- (0.07371635501898201,0.07371635501898201)
(11,0.21687458139534874) +- (0.13127835175540212,0.13127835175540212)
(12,0.14248305813953496) +- (0.05910238846940188,0.05910238846940188)
         };

             \addplot+[
        color=orange, 
           mark size=0.8,
        error bars/.cd,y dir=both,y explicit
        ]
        coordinates {
(1,3.7475593023255827) +- (0.4709899952890919,0.4709899952890919)
(2,2.4030424418604652) +- (0.4314612882370471,0.4314612882370471)
(3,1.296241162790698) +- (0.3750547548167088,0.3750547548167088)
(4,0.8848231395348838) +- (0.260015481203953,0.260015481203953)
(5,0.7013947674418604) +- (0.2609966798656396,0.2609966798656396)
(6,0.35677941860465123) +- (0.18114616352493354,0.18114616352493354)
(7,0.32809177906976733) +- (0.16903988656159932,0.16903988656159932)
(8,0.17492060465116277) +- (0.12926406148004607,0.12926406148004607)
(9,0.13585149999999996) +- (0.08809516119872843,0.08809516119872843)
(10,0.11344482558139539) +- (0.046845373739096155,0.046845373739096155)
(11,0.1372547093023256) +- (0.07458108800988345,0.07458108800988345)
(12,0.07192815116279068) +- (0.026572080005177363,0.026572080005177363)

        }; 
     
    \addplot+[
            color=gray, 
            mark size=0.8,
            error bars/.cd,y dir=both,y explicit
            ]
            coordinates {
(1,3.783689302325582) +- (0.44427950738581934,0.44427950738581934)
(2,2.9385453488372097) +- (0.45629069042006554,0.45629069042006554)
(3,2.448713953488371) +- (0.47444161792875306,0.47444161792875306)
(4,2.0297586046511635) +- (0.47373974388838735,0.47373974388838735)
(5,1.612741744186047) +- (0.42939245142672144,0.42939245142672144)
(6,1.3259988372093026) +- (0.4512095670474634,0.4512095670474634)
(7,1.1453491860465115) +- (0.4896386719082348,0.4896386719082348)
(8,0.8170339534883722) +- (0.3390325980676565,0.3390325980676565)
(9,0.5960024534883724) +- (0.2783122434181672,0.2783122434181672)
(10,0.4071366860465115) +- (0.24148343015797313,0.24148343015797313)
(11,0.23390584883720922) +- (0.127391885640238,0.127391885640238)
(12,0.23703033255813957) +- (0.14344728169486126,0.14344728169486126)

             };

             \addlegendentry{\small ActiveVaR}      
             \addlegendentry{\small ARC}
             \addlegendentry{\small Random}
            % \addlegendentry{AS-Greedy}
        \end{axis}

        \end{tikzpicture}

\caption{\small Averaged policy losses}
\end{subfigure}
~
\begin{subfigure}[b]{0.45\linewidth}
\vspace{-0.5cm}
%\begin{table}[h]
\begin{center}
  \begin{tabular}{ | c | c | }
    \hline
     \textbf{Algorithm} & \textbf{Avg. Time (s)}  \\ \hline
     Random & 0.0015 \\ \hline
     ActiveVaR & 0.0101 \\ \hline
   %  AS-Greedy & 0.4258 \\ \hline
     ARC & 865.6993 \\ 
    \hline
  \end{tabular}
\end{center}
\vspace{1.2cm}
\caption{\small Timing for one iteration of each algorithm}
%\end{table}
\end{subfigure}
\caption{\small Active critique queries in 8$\times$8 gridworlds with 48 features.}
\label{fig:traj_query}
\end{figure}

\subsection{Active Imitation Learning for a 2D Highway Driving Task}
\begin{figure}
\centering
\begin{subfigure}[b]{0.32\linewidth}
\centering
\includegraphics[scale=0.1]{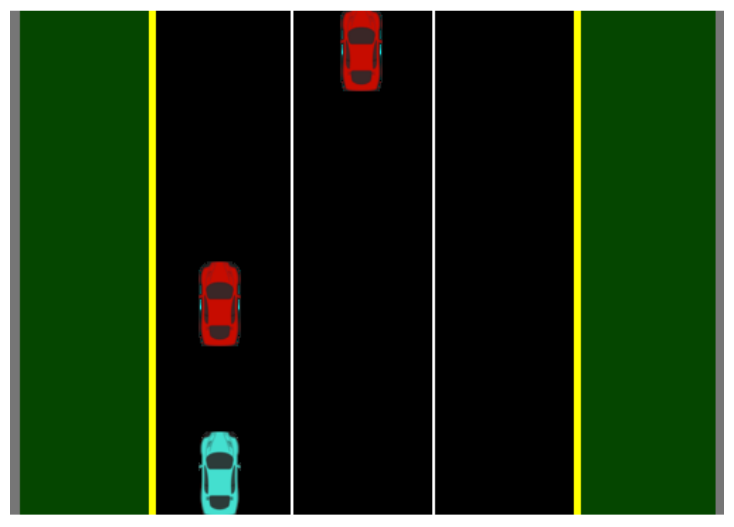}
\includegraphics[scale=0.1]{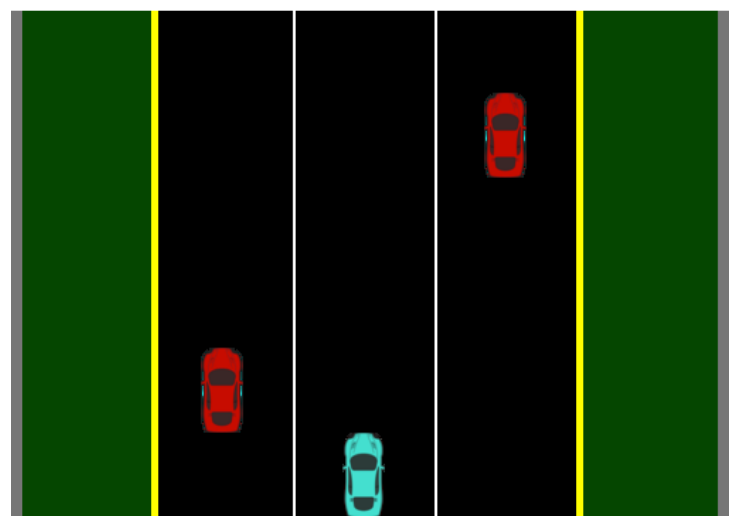} 
\caption{\small 5 Steps of Demos}
\end{subfigure}
\begin{subfigure}[b]{0.32\linewidth}
\centering
\includegraphics[scale=0.1]{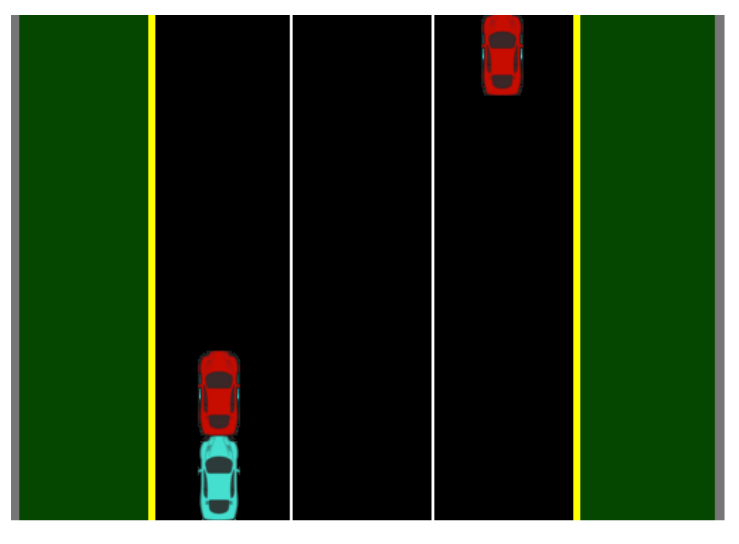}
\includegraphics[scale=0.1]{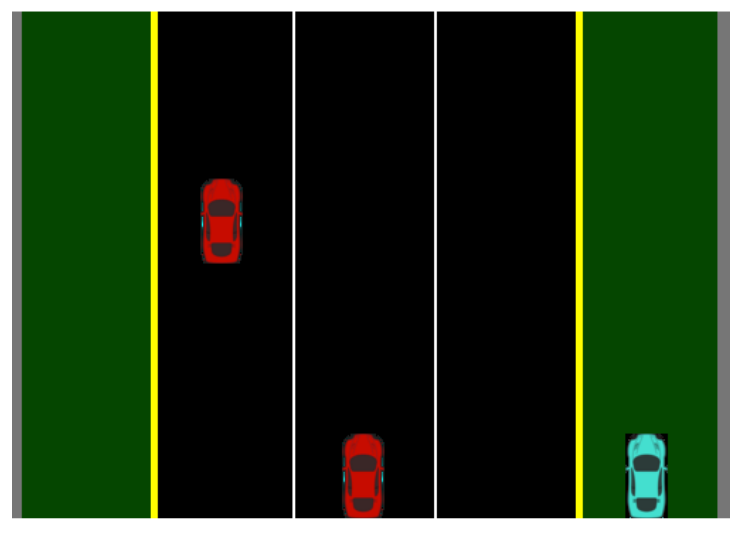}
\caption{\small 10 Steps of Demos}
\end{subfigure}
\begin{subfigure}[b]{0.32\linewidth}
\centering
\includegraphics[scale=0.1]{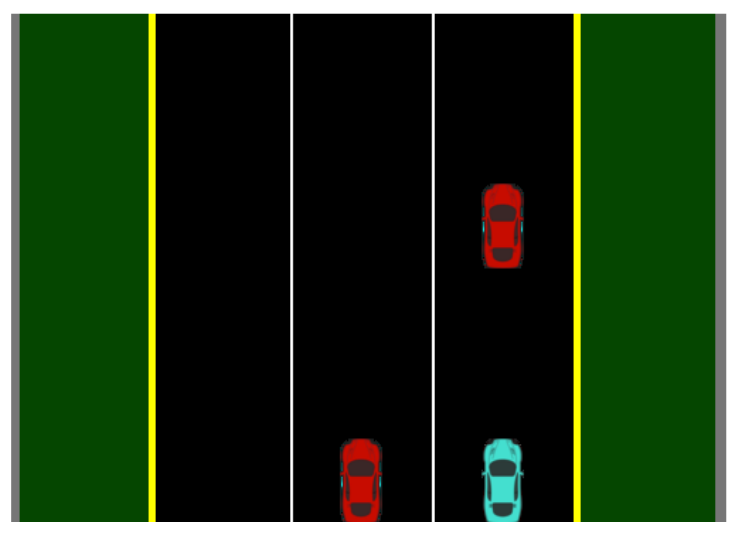}
\includegraphics[scale=0.1]{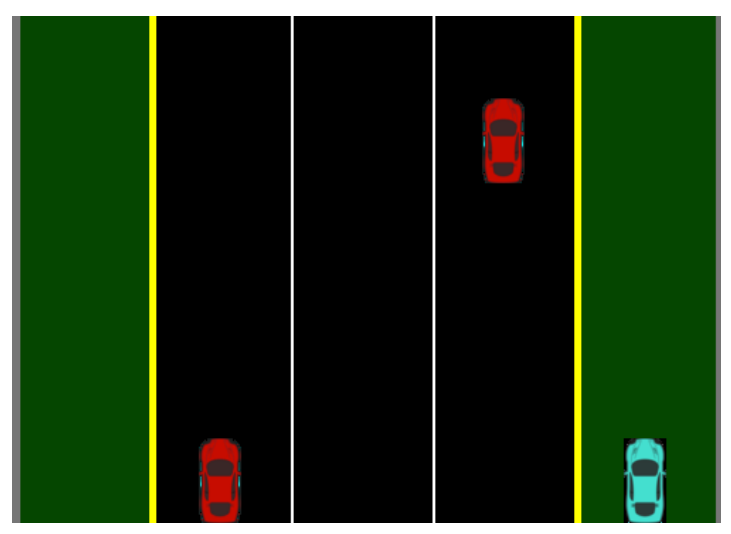}
\caption{\small 20 Steps of Demos}
\end{subfigure}
\caption{\small Active action queries in a 2D highway driving task after different numbers of initial human demonstrations. Initial states are randomly sampled and evaluated; high risk states are selected as active action queries.}
\label{fig:car_sim}
\end{figure}

We also evaluated ActiveVaR for learning from demonstration in a 2D driving simulator, in which the transition dynamics are not deterministic and no ground truth reward function is specified. Human demonstrations are provided by controlling the agent (ourselves) via keyboard. In the 3-lane driving simulator, the controlled agent moves forward two times faster than the two other agents on the road and has three available actions at any given time: moving left, move right, or stay. The states are approximated using discrete features including lane positions and distances to other agents.
Figure~\ref{fig:car_sim} shows the generated active action queries (high-0.95-VaR states) after varying amounts of initial demonstrations are provided. The provided human trajectories demonstrate a safe driving style by avoiding collisions and staying on the road (black lanes). As increasing amounts of initial demonstrations are provided, the active queries start to explore less-frequent states and corner cases that were not addressed in the initial training data.

\subsection{Robot Table Setting Task}
\begin{figure}
\centering
\begin{subfigure}[b]{0.35\linewidth}
\centering
\includegraphics[scale=0.13]{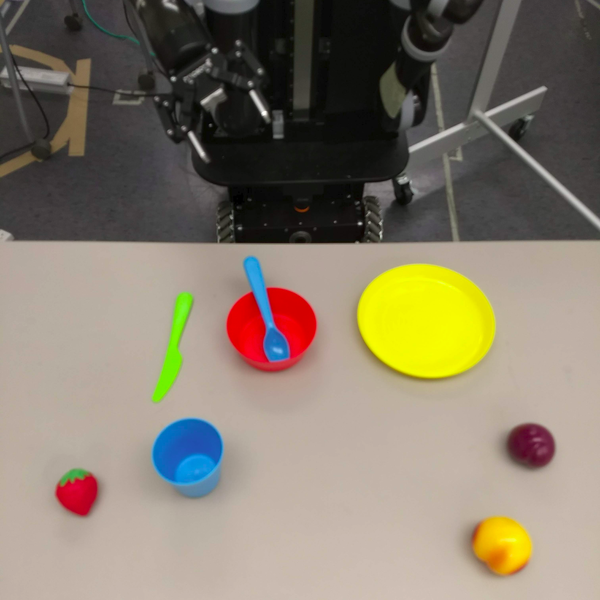}
\caption{\small Place Spoon}
\label{}
\end{subfigure}
~
\begin{subfigure}[b]{0.35\linewidth}
\centering
\includegraphics[scale=0.13]{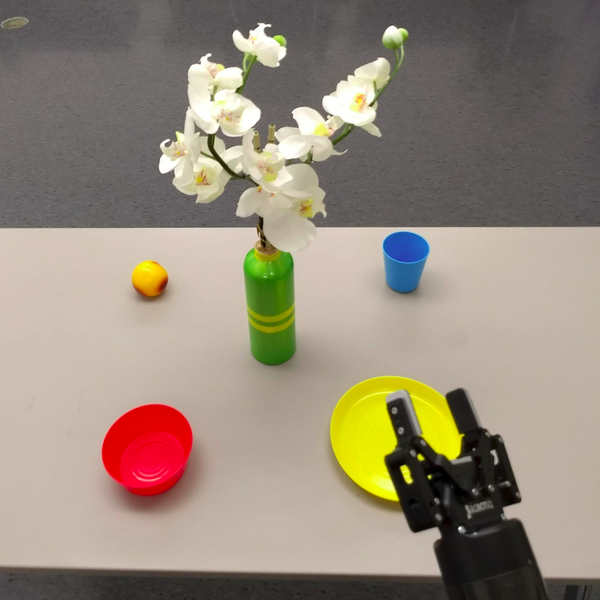}
\caption{\small Place Vase}
\label{}
\end{subfigure}
\caption{\small Setting the table task. (a) Robot actively requests demonstration learning preferences for (a) placing a spoon in the bowl and (b) placing the flower vase in the center of the table.}
\label{fig:robotPlacement}
\vspace{-0.3cm}
\end{figure}

Finally, we consider a robot learning to set a table based on a demonstrator's preferences (see Figure~\ref{fig:robotPlacement}).
We model the reward function as a linear combination of Gaussian radial basis functions. 
Given $k$ items on the table, we assume the reward for placement location $x$ is given by
\begin{equation}
R(x) = \sum_{i=1}^k w_i \cdot \text{rbf}(x,c_i,\sigma^2_i) 
\end{equation}
where
%\begin{equation}
$\text{rbf}(x,c,\sigma^2) = \exp(-\|x-c\|^2/\sigma^2)$.
%\end{equation} 
%Each radial basis function center, $c$, is associated with an object on the table and has a fixed width, $\sigma^2$. 
%The goal is to learn the weights corresponding to the human table placement preferences. 

The demonstrator first gives a demonstration from an initial configuration $C_0$. The robot then needs to infer the correct reward function that matches the demonstrator's intention. We consider an active approach where the robot can generate a novel configuration $C_i$ and ask the demonstrator where it should place the item. The robot hypothesizes multiple configurations $C_i$ and picks the configuration $C^*$ that has maximum 0.95-VaR over its current best guess of the demonstrator's policy. 

% To calculate the VaR we need a distribution over likely reward functions given the demonstrations $D$. We could use a softmax likelihood like BIRL, but this requires computing the partition function over all actions and we assume an infinite action space. Instead, we choose to use the likelihood used by \citet{michini2015bayesian}:
% \begin{equation}
% P(D|R) = \prod_{i=1}^{|D|} \pi(a_i|s_i) = \prod_{i=1}^{|D|}\frac{1}{Z_i} \exp(c Q^*(s,a)) 	
% \end{equation}
% where $\beta$ is a hyperparameter and $a^*_R$ is the optimal action under reward function $R$. \scott{but what is this likelihood and why is it good?} 

% In this case, the states are table configurations and the actions are positions. Thus, the likelihood can be written as 
% \begin{equation}
% P(D|R) = \prod_{i=1}^m \pi(x_i|C_i) = \prod_{i=1}^m \exp(-\beta \|x_i - x^*_R\|^2_2).	
% \end{equation}
% To compute this likelihood, we need to be able to calculate the optimal placement location $x^*_R$ for a radial basis function $R$.

%\subsection{Calculating the best placement}
Given an RBF reward function, the robot needs to estimate an optimal placement position. To calculate the optimal position we use gradient ascent with random restarts and pick the best position. 
 The gradient of the reward with respect to the position $x$ is given by:
 \begin{equation}
 \nabla_x R(x) = \sum_{i=1}^k w_i \cdot \text{rbf}(x,c_i,\sigma^2_i) \bigg( \frac{-2x + 2c_i}{\sigma^2_i}\bigg).
 \end{equation}

%\subsection{Estimating posterior and MAP}
MCMC is used to estimate the posterior $P(R|D)$ and determine the MAP reward $R_{\rm MAP}$, which is used as the best guess of the demonstrator's intent. 
%The MAP reward is the reward function the robot would use at test time so we now want to evaluate our uncertainty over this map reward and corresponding policy and ask for help where we might be uncertain about our performance.
%\subsection{Finding the best active query}
%We measure uncertainty using VaR since it allows us to reason about worst-case outcomes and ask for help in situations where our best guess of the demonstrators intent could result in really bad performance. 
Given the MCMC estimate of the reward posterior, the  robot samples random table configurations $C_j$. For each random configuration, the robot computes the $\alpha$-VaR by first finding the best placement position $x^*_{\rm MAP}$ given by the MAP reward function, and then evaluating this placement position over the estimated posterior found using MCMC. This is done by calculating the placement loss,
$\text{Loss}_i = \|x^i_{R_i} - x^*_{\rm MAP}\|_2, \;\;\forall \; R_i \in P(R|D)$,
and then sorting to estimate the $\alpha$-VaR of that configuration.
%\begin{equation}
%\alpha-\text{VaR}(C_j) = \alpha\text{-quantile} ( %\text{sort}(\text{Loss})).
%\end{equation}
The robot then picks as the query configuration, $C_{\rm query} = \arg \max_j \alpha$-VaR$(C_j)$, and actively asks for a demonstration in this configuration. Note that in this task the epsilon stopping condition can be defined in terms of placement error (distance between where demonstrator would place object and where robot would place object).
%Giving queries based on VaR allows the robot to assess what it knows and does not know, in order to pinpoint configurations that are adversarial for its current best guess of the demonstrator's intention.

%\textbf{In this sense it is kind of like the Abbeel Ng IRL algorithm. Need to think more about this and whether there is a deeper connection between their work and the VaR work...}

\begin{figure}[t] 
\vspace{-0.5cm}
\centering
\begin{subfigure}[b]{0.45\linewidth}
\centering
    \begin{tikzpicture}    
        \begin{axis}[
            name=plot1,
            xlabel={\small Number of Demonstrations},
            x label style={at={(axis description cs:0.5,0.04)},anchor=north},
            ylabel={\small Mean Placement Error},
            axis x line=bottom,
            axis y line=left,
            yticklabel style={/pgf/number format/fixed, /pgf/number format/precision=5		},
            %tick = {0.04,0.08,0.16},
            width=5.5cm,height=4cm,
            xmin=0, xmax=10,
            ymin=0, ymax=0.18,
            ymajorgrids=true,
            xmajorgrids=true,
            xmajorgrids=true,
            grid style=dashed,
            legend style={at={(0.,1.1)},anchor=west},
            legend columns=2,
            legend style={draw=none},
        ]
             \addplot+[
            color=red, 
               mark size=0.8,
            error bars/.cd,y dir=both,y explicit
            ]
            coordinates {
(1, 0.15701566808016101) +- (0.016394096800845335, 0.016394096800845335)
(2, 0.13701288736515199) +- (0.015599874491406554, 0.015599874491406554)
(3, 0.12184647464421895) +- (0.016483833146990208, 0.016483833146990208)
(4, 0.099124446638242034) +- (0.014666294531371169, 0.014666294531371169)
(5, 0.079457264259469029) +- (0.012397394043291583, 0.012397394043291583)
(6, 0.067663243012780025) +- (0.010718650560180128, 0.010718650560180128)
(7, 0.061769012627703998) +- (0.0097247959987700923, 0.0097247959987700923)
(8, 0.05635624454808999) +- (0.0077566066560480207, 0.0077566066560480207)
(9, 0.051358028897801986) +- (0.0068114448586575208, 0.0068114448586575208)
(10, 0.04923607056233003) +- (0.0062149851004561479, 0.0062149851004561479)
(11, 0.044247713083208989) +- (0.0057854104054690511, 0.0057854104054690511)

            };
            
    \addplot+[
            color=orange, 
            mark size=0.8,
            error bars/.cd,y dir=both,y explicit
            ]
            coordinates {
(1, 0.15701566808016101) +- (0.016394096800845335, 0.016394096800845335)
(2, 0.14311313984148497) +- (0.017723674986944968, 0.017723674986944968)
(3, 0.12539644437318803) +- (0.018314822157551477, 0.018314822157551477)
(4, 0.10933938454818398) +- (0.016293878053184097, 0.016293878053184097)
(5, 0.096260347279796002) +- (0.015181259159132675, 0.015181259159132675)
(6, 0.095972098597333022) +- (0.015540154390053953, 0.015540154390053953)
(7, 0.082883836901750996) +- (0.014359208930858104, 0.014359208930858104)
(8, 0.08255963594394404) +- (0.013209597988733652, 0.013209597988733652)
(9, 0.077265800004247001) +- (0.015260974224548068, 0.015260974224548068)
(10, 0.068803436414835009) +- (0.011863356262185522, 0.011863356262185522)
(11, 0.06350645971612498) +- (0.0097692205824695694, 0.0097692205824695694)
       };

          \addlegendentry{ActiveVaR}   
         
          \addlegendentry{Random}
                \addlegendentry{AS} 
\end{axis}
\end{tikzpicture}

\caption{\small Place Vase}
\end{subfigure}
~
\begin{subfigure}[b]{0.45\linewidth}
\centering
    \begin{tikzpicture}    
        \begin{axis}[
            name=plot1,
            xlabel={\small Number of Demonstrations},
            x label style={at={(axis description cs:0.5,0.04)},anchor=north},
            ylabel={\small Max Placement Error},
             y label style={at={(axis description cs:0.0,0.5)},anchor=north},
            axis x line=bottom,
            axis y line=left,
            yticklabel style={/pgf/number format/fixed, /pgf/number format/precision=2		},
            %tick = {0.04,0.08,0.16},
            width=5.5cm,height=4cm,
            xmin=0, xmax=10,
            ymin=0, ymax=0.6,
            ymajorgrids=true,
            xmajorgrids=true,
            xmajorgrids=true,
            grid style=dashed,
            legend style={at={(0.,1.1)},anchor=west},
            legend columns=2,
            legend style={draw=none},
        ]
             \addplot+[
            color=red, 
               mark size=0.8,
            error bars/.cd,y dir=both,y explicit
            ]
            coordinates {
(1, 0.35036146184926514) +- (0.04097853858552776, 0.04097853858552776)
(2, 0.30215580663316999) +- (0.03941146752787221, 0.03941146752787221)
(3, 0.26094141144388994) +- (0.037751379673464418, 0.037751379673464418)
(4, 0.2102043452361379) +- (0.031559048673852209, 0.031559048673852209)
(5, 0.16927683612445002) +- (0.026864470317134855, 0.026864470317134855)
(6, 0.14335898233112604) +- (0.022315310826339699, 0.022315310826339699)
(7, 0.12910354088132994) +- (0.020851263174230156, 0.020851263174230156)
(8, 0.11953855742248601) +- (0.014999488301026717, 0.014999488301026717)
(9, 0.10965842845862099) +- (0.014123851473302449, 0.014123851473302449)
(10, 0.10304855502299105) +- (0.012116660837833118, 0.012116660837833118)
(11, 0.093338807988088024) +- (0.011487400395275384, 0.011487400395275384)

            };
            
    \addplot+[
            color=black, 
            mark size=0.8,
            error bars/.cd,y dir=both,y explicit
            ]
            coordinates {
(1, 0.57551865390222989) +- (0.021482712402438087, 0.021482712402438087)
(2, 0.50182926825355989) +- (0.029246055525336469, 0.029246055525336469)
(3, 0.45881288912072987) +- (0.038670984591226136, 0.038670984591226136)
(4, 0.39822254691557007) +- (0.0367588450521305, 0.0367588450521305)
(5, 0.31302808965705003) +- (0.034476798052786826, 0.034476798052786826)
(6, 0.24755863170201983) +- (0.028433877037246227, 0.028433877037246227)
(7, 0.21373882833303001) +- (0.025451760743408582, 0.025451760743408582)
(8, 0.19479207137311999) +- (0.016983573943692324, 0.016983573943692324)
(9, 0.17457778074495994) +- (0.014919281760211289, 0.014919281760211289)
(10, 0.16458401041747003) +- (0.013753140722197972, 0.013753140722197972)

       };

          \addlegendentry{\small Actual}   
         
          \addlegendentry{\small Upper bound}
                
\end{axis}
\end{tikzpicture}

\caption{\small Place Vase}
\end{subfigure}
%\caption{Learning to place a flower vase in the center of a cluttered table. (a) Comparison of active verus random queries for learning  (b) The 0.95-VaR placement error bound provides an upper bound on the actual maximum placement error. Results are averaged from 100 trials with 0.5 standard deviation error bars.. Placement error is calculated using 200 random test configurations. \textbf{TODO: still waiting for some of the jobs to finish on condor} }
%\end{figure}

%\begin{figure}[h] 
%\label{fig:place_flowers}
%\centering
\begin{subfigure}[b]{0.45\linewidth}
\centering
    \begin{tikzpicture}    
        \begin{axis}[
            name=plot1,
            xlabel={\small Number of Demonstrations},
            x label style={at={(axis description cs:0.5,0.04)},anchor=north},
            ylabel={\small Mean Placement Error},
            axis x line=bottom,
            axis y line=left,
            yticklabel style={/pgf/number format/fixed, /pgf/number format/precision=5		},
            %tick = {0.04,0.08,0.16},
            width=5.5cm,height=4cm,
            xmin=0, xmax=10,
            ymin=0, ymax=0.35,
            ymajorgrids=true,
            xmajorgrids=true,
            xmajorgrids=true,
            grid style=dashed,
            legend style={at={(0.,1.1)},anchor=west},
            legend columns=2,
            legend style={draw=none},
        ]
             \addplot+[
            color=red, 
               mark size=0.8,
            error bars/.cd,y dir=both,y explicit
            ]
            coordinates {
(1, 0.29504440696059886) +- (0.058108413071385465, 0.058108413071385465)
(2, 0.21543145327613689) +- (0.066680041454668937, 0.066680041454668937)
(3, 0.15574120765992697) +- (0.052537895608921262, 0.052537895608921262)
(4, 0.12310380153548998) +- (0.050302735650850541, 0.050302735650850541)
(5, 0.090043781138633983) +- (0.030392213229864323, 0.030392213229864323)
(6, 0.075645820040550996) +- (0.016361399006735357, 0.016361399006735357)
(7, 0.067791576303663018) +- (0.013911317965297522, 0.013911317965297522)
(8, 0.057806120378348015) +- (0.01034642297683507, 0.01034642297683507)
(9, 0.052461790890379986) +- (0.009771670791335247, 0.009771670791335247)
(10, 0.048084978893326004) +- (0.011456742773943493, 0.011456742773943493)
(11, 0.041596631224087986) +- (0.0057850686564742239, 0.0057850686564742239)

            };
            
    \addplot+[
            color=orange, 
            mark size=0.8,
            error bars/.cd,y dir=both,y explicit
            ]
            coordinates {
(1, 0.29504440696059886) +- (0.058108413071385465, 0.058108413071385465)
(2, 0.20740147063919495) +- (0.057480027978405936, 0.057480027978405936)
(3, 0.190857356081227) +- (0.059585906484185616, 0.059585906484185616)
(4, 0.14949429780681503) +- (0.048319876362444607, 0.048319876362444607)
(5, 0.14452009678807201) +- (0.054955969816588977, 0.054955969816588977)
(6, 0.14617627653380003) +- (0.056104625596259013, 0.056104625596259013)
(7, 0.13214801564139503) +- (0.052113227196500481, 0.052113227196500481)
(8, 0.10403891139001999) +- (0.046716140098749449, 0.046716140098749449)
(9, 0.089879308281127987) +- (0.03921950734017815, 0.03921950734017815)
(10, 0.08643233380390701) +- (0.039075064466795192, 0.039075064466795192)
(11, 0.065194844811520006) +- (0.025061409999892998, 0.025061409999892998)

       };

          \addlegendentry{\small ActiveVaR}   
         
          \addlegendentry{\small Random}
\end{axis}
\end{tikzpicture}

\caption{\small Place Spoon}
\end{subfigure}
~
\begin{subfigure}[b]{0.45\linewidth}
\centering
    \begin{tikzpicture}    
        \begin{axis}[
            name=plot1,
            xlabel={\small Number of Demonstrations},
            x label style={at={(axis description cs:0.5,0.04)},anchor=north},
            ylabel={\small Max Placement Error},
            y label style={at={(axis description cs:0.0,0.5)},anchor=north},
            axis x line=bottom,
            axis y line=left,
            yticklabel style={/pgf/number format/fixed, /pgf/number format/precision=2		},
            %tick = {0.04,0.08,0.16},
            width=5.5cm,height=4cm,
            xmin=0, xmax=10,
            ymin=0, ymax=1.2,
            ymajorgrids=true,
            xmajorgrids=true,
            xmajorgrids=true,
            grid style=dashed,
            legend style={at={(0.,1.1)},anchor=west},
            legend columns=2,
            legend style={draw=none},
        ]
             \addplot+[
            color=red, 
               mark size=0.8,
            error bars/.cd,y dir=both,y explicit
            ]
            coordinates {
(1, 0.91608506715169025) +- (0.13946487881415334, 0.13946487881415334)
(2, 0.75086869309845272) +- (0.1841754930563001, 0.1841754930563001)
(3, 0.59480130315717972) +- (0.19418089445916245, 0.19418089445916245)
(4, 0.4225314269340999) +- (0.16720463849802678, 0.16720463849802678)
(5, 0.30513910757867707) +- (0.12840376744790841, 0.12840376744790841)
(6, 0.25048151549791292) +- (0.11270120630475643, 0.11270120630475643)
(7, 0.20879859489197286) +- (0.088208042662659955, 0.088208042662659955)
(8, 0.16847995786072198) +- (0.068808336466269021, 0.068808336466269021)
(9, 0.14538581139802909) +- (0.063975000057558629, 0.063975000057558629)
(10, 0.12653525642492799) +- (0.048258742408446305, 0.048258742408446305)
(11, 0.10329418442047202) +- (0.015143062568312066, 0.015143062568312066)

            };
            
    \addplot+[
            color=black, 
            mark size=0.8,
            error bars/.cd,y dir=both,y explicit
            ]
            coordinates {
(1, 0.96352394964136012) +- (0.10537103871205244, 0.10537103871205244)
(2, 0.97055179778763001) +- (0.10107232628482127, 0.10107232628482127)
(3, 1.0022594980447495) +- (0.095073217845239449, 0.095073217845239449)
(4, 1.0093055110978297) +- (0.095873993762033538, 0.095873993762033538)
(5, 0.99553014350117963) +- (0.094310344600788007, 0.094310344600788007)
(6, 0.9703223081391501) +- (0.09909731298292257, 0.09909731298292257)
(7, 0.87732986446208994) +- (0.13830393336290969, 0.13830393336290969)
(8, 0.75587510869785968) +- (0.16653485335910398, 0.16653485335910398)
(9, 0.69133076031131979) +- (0.18090557001921156, 0.18090557001921156)
(10, 0.55935261176954976) +- (0.17217233159766338, 0.17217233159766338)

       };       
    
          \addlegendentry{\small Actual}   
          \addlegendentry{\small Upper bound}
                
\end{axis}
\end{tikzpicture}

\caption{\small Place Spoon}
\end{subfigure}
\caption{\small Results for learning to place a flower vase and learning to place a spoon on a cluttered table. Active queries in (a) and (c) result in lower error than random queries. The 0.95-VaR placement error bound shown in (b) and (d) provides an upper bound on the actual maximum placement error. All results are averaged from 100 trials with 0.5 standard deviation error bars. Placement error is calculated using 200 random test configurations.}
\label{fig:place_flowers}
\vspace{-0.3cm}
\end{figure}
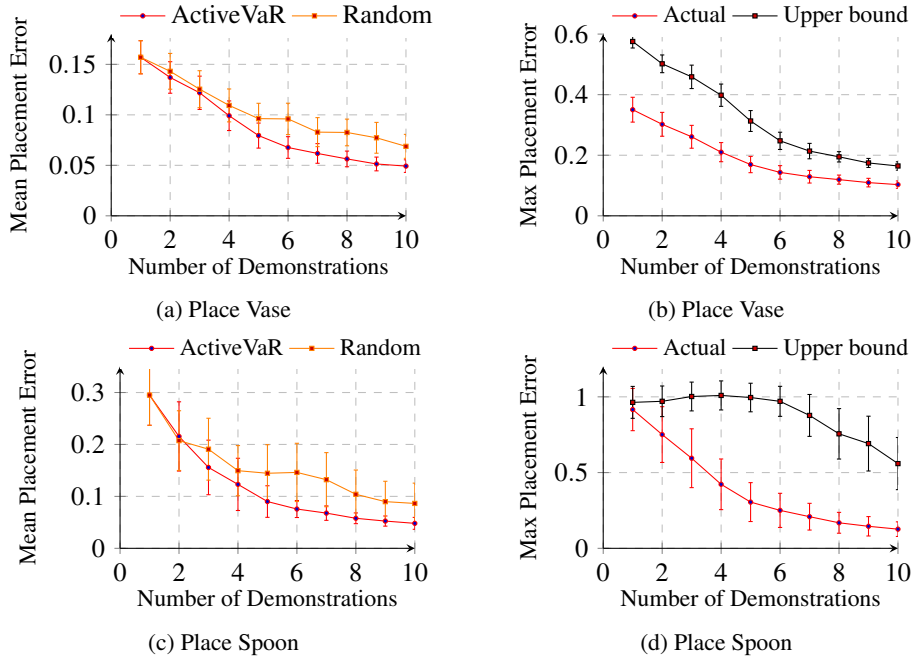

%\subsubsection{Results}
Figure~\ref{fig:place_flowers} shows the results of two table arrangement experiments. In the first experiment the demonstrator teaches the robot to place a vase of flowers on a table with 4 existing objects. The preference is to place the vase in the center of the table, while avoiding placing it on top of other objects. In the second experiment, the demonstrator teaches the robot to place a spoon in a bowl on a table with 6 distractor objects. To allow for expressive reward function hypotheses, we model the reward using RBFs centered on all objects on the table along with 9 fixed RBFs evenly spaced on the table to allow for the  possibility of an absolute placement preference. Possible query configurations are randomly generated by changing the position of one of the objects on the table. We generated synthetic ground truth rewards that corresponded to each placement preference and generated synthetic demonstrations using these ground truth rewards. This allows us to rigorously test the active learning process without needing to physically move the objects for each query configuration. We then validated the learned reward function weights on the real robot in a variety of test configurations (video available at \url{https://youtu.be/yYFae_Obp-0}).

Figures~\ref{fig:place_flowers}(a) and (c) show a comparison of random queries with actively querying the configuration with maximum 0.95-VaR out of 50 randomly generated configurations. Choosing risk-aware queries over random queries results in smaller generalization error. To test whether 0.95-VaR provides a meaningful and accurate upper bound on the true performance of a policy we compared the actual worst-case placement loss under for the robot's current policy after each demonstration with the 0.95-VaR upper bound. Figures~\ref{fig:place_flowers}(b) and (d) demonstrate that the 0.95-VaR upper bound accurately upper bounds the actual worst-case performance and that this bound becomes tighter as more demonstrations are received.

%params
%    num_queries = 10
%    num_test = 100
%    num_configs_halucinate = 30
%    alpha = 0.75  
%    beta=100.0
%    num_steps = 2500
%    step_std = 0.05
%    burn = 0
%    skip = 25
%obj_weights = np.array([-0.1, -0.1, -0.1, -0.1])
%abs_weights = np.array([0.0, 0.0, 0.0, 
%                            0.0, 0.6, 0.0, 
%                            0.0, 0.0, 0.0])

%%===============================================================================
%
%\section{Citations}
%\label{sec:citations}
%
%	Citations can be made using either \textbackslash citep\{\} or \textbackslash citet\{\}, depending from the appropriateness. To avoid the citation moving to the next line, it is often a good practice to replace the space before with a tilde (\~{}) character.
%	Example 1: ``CoRL is the best conference ever, as discussed in~\citep{Calandra2016}.``
%	Example 2: ``\citet{Calandra2016} proved, both theoretically and numerically, that CoRL is the best conference ever.``

%===============================================================================
\section{Conclusion} 
\label{sec:conclusion}

We proposed the first active IRL technique that is based on the actual performance of the policy learned from demonstrations. We compared our approach against existing active IRL algorithms and found that our risk-aware approach outperforms entropy based queries in terms of sample complexity and is comparable to active queries based on information gain while requiring three orders of magnitude less computation on the domains we tested. Experiments in simulated 2-d navigation and highway driving domains, as well as robot table placement tasks, demonstrate that risk-aware active queries allow robots to ask for help in areas of the state-space where the robot has the potential for high generalization error. Our approach allows the robot to upper bound its own policy loss and can be used to let a robot to know when it has received enough demonstrations to safely perform a task.

%===============================================================================

% The maximum paper length is 8 pages excluding references and acknowledgements, and 10 pages including references and acknowledgements

\clearpage
% The acknowledgments are automatically included only in the final version of the paper.
\acknowledgments{This work has taken place in the Personal Autonomous Robotics Lab (PeARL) at The University of Texas at Austin. PeARL research is supported in part by the NSF (IIS-1724157, IIS-1638107, IIS-1617639, IIS-1749204) and ONR (N00014-18-2243).}

%===============================================================================

\bibliography{references}  % .bib

\begin{thebibliography}{22}
\providecommand{\natexlab}[1]{#1}
\providecommand{\url}[1]{\texttt{#1}}
\expandafter\ifx\csname urlstyle\endcsname\relax
  \providecommand{\doi}[1]{doi: #1}\else
  \providecommand{\doi}{doi: \begingroup \urlstyle{rm}\Url}\fi

\bibitem[Argall et~al.(2009)Argall, Chernova, Veloso, and
  Browning]{argall2009survey}
B.~D. Argall, S.~Chernova, M.~Veloso, and B.~Browning.
\newblock A survey of robot learning from demonstration.
\newblock \emph{Robotics and autonomous systems}, 57\penalty0 (5):\penalty0
  469--483, 2009.

\bibitem[Ng et~al.(2000)Ng, Russell, et~al.]{ng2000algorithms}
A.~Y. Ng, S.~J. Russell, et~al.
\newblock Algorithms for inverse reinforcement learning.
\newblock In \emph{Icml}, pages 663--670, 2000.

\bibitem[Abbeel and Ng(2004)]{abbeel2004apprenticeship}
P.~Abbeel and A.~Y. Ng.
\newblock Apprenticeship learning via inverse reinforcement learning.
\newblock In \emph{Proceedings of the twenty-first international conference on
  Machine learning}, page~1. ACM, 2004.

\bibitem[Sutton and Barto(1998)]{sutton1998reinforcement}
R.~S. Sutton and A.~G. Barto.
\newblock \emph{Reinforcement learning: An introduction}, volume~1.
\newblock MIT press Cambridge, 1998.

\bibitem[Lopes et~al.(2009)Lopes, Melo, and Montesano]{lopes2009active}
M.~Lopes, F.~Melo, and L.~Montesano.
\newblock Active learning for reward estimation in inverse reinforcement
  learning.
\newblock In \emph{Joint European Conference on Machine Learning and Knowledge
  Discovery in Databases}, pages 31--46. Springer, 2009.

\bibitem[Cohn et~al.(2011)Cohn, Durfee, and Singh]{cohn2011comparing}
R.~Cohn, E.~Durfee, and S.~Singh.
\newblock Comparing action-query strategies in semi-autonomous agents.
\newblock In \emph{The 10th International Conference on Autonomous Agents and
  Multiagent Systems-Volume 3}, pages 1287--1288. International Foundation for
  Autonomous Agents and Multiagent Systems, 2011.

\bibitem[Cui and Niekum(2018)]{cui2018active}
Y.~Cui and S.~Niekum.
\newblock Active reward learning from critiques.
\newblock In \emph{IEEE International Conference on Robotics and Automation
  (ICRA)}. IEEE, 2018.

\bibitem[Amodei et~al.(2016)Amodei, Olah, Steinhardt, Christiano, Schulman, and
  Man{\'e}]{amodei2016concrete}
D.~Amodei, C.~Olah, J.~Steinhardt, P.~Christiano, J.~Schulman, and D.~Man{\'e}.
\newblock Concrete problems in ai safety.
\newblock \emph{arXiv preprint arXiv:1606.06565}, 2016.

\bibitem[Thomas et~al.(2017)Thomas, da~Silva, Barto, and
  Brunskill]{thomas2017ensuring}
P.~S. Thomas, B.~C. da~Silva, A.~G. Barto, and E.~Brunskill.
\newblock On ensuring that intelligent machines are well-behaved.
\newblock \emph{arXiv preprint arXiv:1708.05448}, 2017.

\bibitem[Chow et~al.(2015)Chow, Tamar, Mannor, and Pavone]{chow2015risk}
Y.~Chow, A.~Tamar, S.~Mannor, and M.~Pavone.
\newblock Risk-sensitive and robust decision-making: a cvar optimization
  approach.
\newblock In C.~Cortes, N.~D. Lawrence, D.~D. Lee, M.~Sugiyama, and R.~Garnett,
  editors, \emph{Advances in Neural Information Processing Systems 28}, pages
  1522--1530. 2015.

\bibitem[Brown et~al.(2017)Brown, Hudack, Gemelli, and
  Banerjee]{brown2017exact}
D.~S. Brown, J.~Hudack, N.~Gemelli, and B.~Banerjee.
\newblock Exact and heuristic algorithms for risk-aware stochastic physical
  search.
\newblock \emph{Computational Intelligence}, 33\penalty0 (3):\penalty0
  524--553, 2017.

\bibitem[Tamar et~al.(2015)Tamar, Glassner, and Mannor]{tamar2015optimizing}
A.~Tamar, Y.~Glassner, and S.~Mannor.
\newblock Optimizing the cvar via sampling.
\newblock In \emph{Twenty-Ninth AAAI Conference on Artificial Intelligence},
  pages 2993--2999, 2015.

\bibitem[Garc{\i}a and Fern{\'a}ndez(2015)]{garcia2015comprehensive}
J.~Garc{\i}a and F.~Fern{\'a}ndez.
\newblock A comprehensive survey on safe reinforcement learning.
\newblock \emph{Journal of Machine Learning Research}, 16\penalty0
  (1):\penalty0 1437--1480, 2015.

\bibitem[Brown and Niekum(2018)]{brown2018efficient}
D.~S. Brown and S.~Niekum.
\newblock Efficient probabilistic performance bounds for inverse reinforcement
  learning.
\newblock In \emph{AAAI Conference on Artificial Intelligence}, 2018.

\bibitem[Knox et~al.(2013)Knox, Stone, and Breazeal]{knox2013teaching}
W.~B. Knox, P.~Stone, and C.~Breazeal.
\newblock Teaching agents with human feedback: a demonstration of the tamer
  framework.
\newblock In \emph{Proceedings of the companion publication of the 2013
  international conference on Intelligent user interfaces companion}, pages
  65--66. ACM, 2013.

\bibitem[MacGlashan et~al.(2017)MacGlashan, Ho, Loftin, Peng, Roberts, Taylor,
  and Littman]{macglashan2017interactive}
J.~MacGlashan, M.~K. Ho, R.~Loftin, B.~Peng, D.~Roberts, M.~E. Taylor, and
  M.~L. Littman.
\newblock Interactive learning from policy-dependent human feedback.
\newblock \emph{arXiv preprint arXiv:1701.06049}, 2017.

\bibitem[Sadigh et~al.(2017)Sadigh, Dragan, Sastry, and
  Seshia]{sadigh2017active}
D.~Sadigh, A.~Dragan, S.~S. Sastry, and S.~A. Seshia.
\newblock Active preference-based learning of reward functions.
\newblock In \emph{Proceedings of Robotics: Science and Systems (RSS)}, page~1,
  2017.

\bibitem[Christiano et~al.(2017)Christiano, Leike, Brown, Martic, Legg, and
  Amodei]{christiano2017deep}
P.~Christiano, J.~Leike, T.~B. Brown, M.~Martic, S.~Legg, and D.~Amodei.
\newblock Deep reinforcement learning from human preferences.
\newblock \emph{arXiv preprint arXiv:1706.03741}, 2017.

\bibitem[Syed and Schapire(2008)]{syed2008game}
U.~Syed and R.~E. Schapire.
\newblock A game-theoretic approach to apprenticeship learning.
\newblock In \emph{Advances in neural information processing systems}, pages
  1449--1456, 2008.

\bibitem[Brown and Niekum(2017)]{brown2017toward}
D.~S. Brown and S.~Niekum.
\newblock Toward probabilistic safety bounds for robot learning from
  demonstration.
\newblock In \emph{AAAI Fall Symposium on AI for HRI}, 2017.

\bibitem[Jorion(1997)]{jorion1997value}
P.~Jorion.
\newblock \emph{Value at risk: the new benchmark for controlling market risk}.
\newblock Irwin Professional Pub., 1997.

\bibitem[Ramachandran and Amir(2007)]{ramachandran2007bayesian}
D.~Ramachandran and E.~Amir.
\newblock Bayesian inverse reinforcement learning.
\newblock In \emph{Proc. of 20th International Joint Conference of Artificial
  Intelligence, 2007}, pages 2586--2591, 2007.

\end{thebibliography}
\clearpage

\begin{appendices}
\section{Additional Results: Comparing ActiveVaR and Random }
\label{app:results}
In the experiments of navigation in random gridworld (section 5.3), as shown in Figure~\ref{fig:traj_query}, the improvement of ActiveVaR over Random is not prominent. We believe it is due to the fact that we used dense features and dense rewards that make random queries informative. Therefore, we ran additional experiments where we force the true reward and features to be sparse and the gridworld has only a few informative initial states such that there is a lower chance to sample a trajectory from an informative state. 

\begin{figure}[h]
\centering
\begin{subfigure}[b]{0.45\linewidth}
\centering
\includegraphics[scale=0.38]{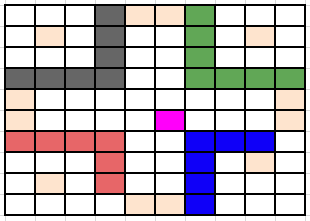}
\caption{\small Gridworld Setup: each color (except light orange) is an unique feature; pink feature is the only feature with positive weight, other colors all have negative weights; states shaded with light orange are designated initial states.}
\end{subfigure}
\begin{subfigure}[b]{0.45\linewidth}
\centering
\includegraphics[scale=0.2]{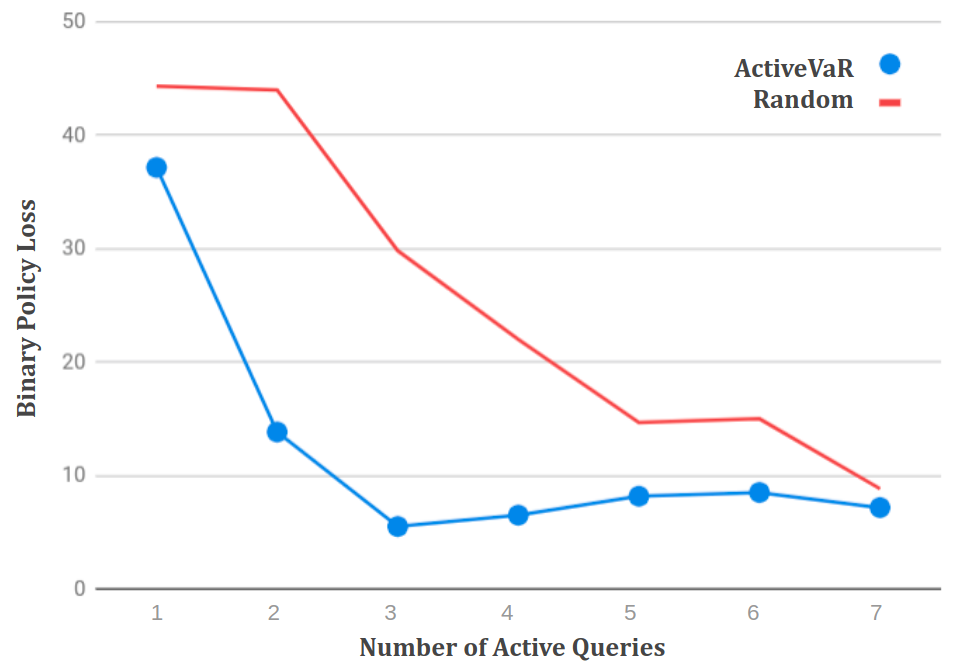}
\caption{\small Binary Policy Loss Over Queries}
\end{subfigure}

\caption{\small Example Gridworld Navigation Experiment with Sparse Feature and Reward.}
\label{fig:new_exp}
\end{figure}

Figure~\ref{fig:new_exp} shows a selected setup and the policy loss averaged over 10 different runs in the selected environment. As shown in the plot, under this setting, ActiveVaR has a much larger improvement over random. 

We also computed the worst-case actual policy loss, as this is what our method seeks to minimize, on a similar barrier domain with all possible states as initial states. In this setting random queries require on average 3.45 times more demonstrations than activeVaR queries to achieve low $(<0.01)$ worst-case policy loss. 

\section{Choosing an Intuitive Stopping Condition}
\label{app:discussion}
The $\epsilon$ stopping criterion in our work is based on an upper bound on performance loss. We know of no other work that has proposed or used such a stopping condition. Without context, an $\epsilon$ stopping condition based on raw policy loss may not be easy to select; however, policy loss can be normalized to obtain a percentage that is more semantically meaningful. The normalization can be computed as 
\begin{equation} 
\text{normalized EVD}(s,R \mid \pi_{\rm MAP}) = \frac{V^{\pi^*}_{R}(s)-V^{\pi_{\text{MAP}}}_{R}(s)}{V^{\pi^*}_{R}(s)}.
\end{equation}
Using the normalized EVD in place of EVD in the above algorithms allows us to calculate an upper bound on the normalized Value-at-Risk.

For example, if the normalized $\alpha$-VaR is 5\% then we can say with high-confidence that the expected return of the learned policy is within 5\% of the expected return of the optimal policy under the demonstrators reward. This allows epsilon to be set as a fixed percentage such as 5\% of 1\% depending on risk aversion.

\end{appendices}

\end{document}